\documentclass[10pt,twocolumn,letterpaper]{article}

\usepackage{cvpr}
\usepackage{times}
\usepackage{epsfig}
\usepackage{graphicx}
\usepackage{amsmath}
\usepackage{amssymb}
\usepackage{enumitem}

\usepackage[breaklinks=true,bookmarks=false]{hyperref}

\cvprfinalcopy 

\def\cvprPaperID{****} 

\begin{document}

\title{Block-optimized Variable Bit Rate Neural Image Compression}

\author{Caglar Aytekin, Xingyang Ni, Francesco Cricri, Jani Lainema, Emre Aksu, Miska Hannuksela\\
Nokia Technologies\\
Tampere, Finland\\
{\tt\small \{caglar.aytekin, xingyang.ni.ext, francesco.cricri\}@nokia.com}\\
{\tt\small \{jani.lainema, emre.aksu, miska.hannuksela\}@nokia.com}
}

\maketitle

\begin{abstract}
   In this work, we propose an end-to-end block-based auto-encoder system for image compression. We introduce novel contributions to neural-network based image compression, mainly in achieving binarization simulation, variable bit rates with multiple networks, entropy-friendly representations, inference-stage code optimization and performance-improving normalization layers in the auto-encoder. We evaluate and show the incremental performance increase of each of our contributions. 
   
\end{abstract}

\section{Introduction}
\label{intro}

Image compression has traditionally been addressed by transform-based methods such as JPEG \cite{Wallace1991} and BPG \cite{Sullivan2012}. Recently, neural network based approaches have also been utilized such as hybrid approaches, where neural networks are used together with a traditional codec, or end-to-end learned approaches, where the codec consists solely of neural networks. 

Regarding the hybrid approach, several works involve using neural networks as post-processing filters (\cite{LukasCavigelli2017}, \cite{ChaoDong2015}), to enhance the decoded image. 
In \cite{FengJiang2017} and \cite{Zhao2017} both pre-processing and post-processing neural networks are used. 
In \cite{FengJiang2017} and \cite{Zhao2017}, due to the non-differentiable traditional codec, an end-to-end training cannot be achieved.
\cite{Zhao2017} proposes to utilize alternate training to overcome this issue.
In the first stage, the pre-processing network is trained via a differentiable virtual codec.
In the second stage, the real codec is used and only the post-processing network is trained.

Regarding the end-to-end learned approach, a typical architecture consists of an auto-encoder (see \cite{Rippel2017}, \cite{Theis2017}), where the encoder maps the input image to a low-dimensional tensor, and the decoder reconstructs the image. 

The encoder's output, which typically consists of floating-point values, needs to be quantized in order to achieve reasonable compression rates. 
The quantization operation would provide zero gradients almost everywhere.
In order to approximate the quantization, \cite{JohannesBalle2017} propose to add a random sample from a uniform distribution. 
\cite{Toderici2017} uses a random mapping of floating-point values to binary values with a probability derived from the floating-point value.

Training of auto-encoders for data compression needs to account for both decoding quality and compression efficiency. 
One straightforward training loss for decoding quality is the mean squared error (MSE) between input and output of the auto-encoder. 
Minimizing the MSE maximizes the peak signal to noise ratio (PSNR), which is a widely used evaluation metric in data compression. 
However, a model trained with MSE loss tends to result into blurred decoded images. 
Alternative losses are variational losses \cite{Gregor2016}, adversarial losses \cite{Agustsson2018} and structural similarity loss \cite{Theis2017}.

Regarding the compression efficiency, \cite{Rippel2017} proposes to use an adaptive codelength regularization term which encourages structure in the code, so that the arithmetic coder can exploit it for adapting the final codelength to the complexity of the input. 
In \cite{JohannesBalle2017} and \cite{Theis2017} the authors optimize for rate-distortion performance, where the rate is represented by the entropy. 

Other neural network architectures used for image compression include recurrent models, such as in \cite{Toderici2017}.

In this paper, we propose a system for block-based image compression using auto-encoders. In particular, our contributions are:
\begin{itemize} [noitemsep,wide=0pt, leftmargin=\dimexpr\labelwidth + 2\labelsep\relax]
\item Using multiple networks for variable bit rate, with inference-stage code optimization.
\item Using $L_2$ normalization layer as the first layer of decoder, which improves the training and inference performance.
\item An entropy-friendly loss designed for block-based neural auto-encoders.
\item Fine-tuning each network on a separate sub-set of blocks, according to the blocks' encoding difficulty.
\item Interval-preserving binarization noise, which ensures that the noisy signal is in a certain interval to provide consistent input to the decoder during training.
\end{itemize}

\section{Method}

In this section, we describe the method used in our end-to-end image compression. 
Our method is based on fully-convolutional deep auto-encoders and is applied on 32x32 blocks from the image.


\subsection{Network Description}

\textbf{Auto-Encoder Network}: The encoder part contains five consecutive convolutional blocks.
Each block consists of a convolutional layer with stride 2 followed by a parametric rectified linear unit (PReLU).
These five blocks are followed by a 1x1 convolutional layer and a sigmoid activation. 
The output of this layer is the compressed signal and will be referred to as block-codes from now on.
The block-codes are 1-dimensional, as the input to the network is of size 32x32 and there exits 5 downsampling convolutions.

The first layer of the decoder is an $L_2$ normalization layer.
It has been shown that mapping auto-encoder representations to the hypersphere surface improves clustering \cite{Aytekin2018}.
We also find $L_2$ normalization beneficial for this work, and we will provide more details about the benefits later in the experimental results.
The $L_2$ normalization layer is followed by five consecutive deconvolutional blocks, each upsampling to double size.
Each deconvolutional block consists of a deconvolutional layer followed by PReLU.
The five deconvolutional blocks are followed by a final 1x1 convolutional layer with sigmoid activation.

\textbf{Multiple Networks}: The block-code length (number of vector's entries) for the above network is fixed and equal to the number of convolutional kernels in the last layer of the encoder.
Setting up a fixed code-length for all blocks can be suboptimal, as blocks may have different content complexity and thus different compression difficulty.
To allow for variable bit rate encoding (other than entropy coding), we make use of three separate networks with different code-lengths.
We encode/decode each block with the network that provides the smallest bit rate for a target PSNR value.

\textbf{Deblocking Network}:
Due to block-based compression, the decoded image contains blocking artefacts. 
To suppress these artefacts, we employ a fully-convolutional deblocking filter that operates over the entire image. The network's structure is similar to U-Net \cite{UNET}. 


\subsection{Inference}
During encoding, first the image is divided into 32x32 blocks by raster-scan. 
Each block is encoded by the lowest bit rate neural network (out of three) which satisfies a target PSNR. The output of the encoder is binarized.

We optimize each block-code by optimizing the encoder per block: we keep the weights of the decoder frozen and fine-tune the encoder for a single block.
To this end, we set a target PSNR and start optimizing the encoder of lowest bit rate neural network.
If this network cannot achieve the target PSNR, we move on to the higher bit rate neural network and optimize its encoder.
This process is continued until the target PSNR is achieved and the corresponding block-code is selected as the final one.

We use two-bit indicator signal for each block indicating which neural network was used for encoding that block.
All the indicator signals are concatenated and one long indicator vector is obtained for the entire image. 
The indicator vector is entropy-coded for further bit rate reduction. 
Similarly, each block-code is concatenated to obtain a long image-code. 
This image-code is first difference-coded and then entropy-coded. 
In the end, each image is encoded into three vectors: 1) entropy-coded image-code 2) entropy-coded indicator vector 3) shape of the original image. 

During decoding, first the entropy-coded vectors are decoded. 
Then, the next two bits from indicator vector is read and based on the indicator, the decoder knows which of the three decoder network needs to be used for the current block and therefore the encoding dimension. 
Then the next bit sequence of same length as this encoding dimension is read from the image-code and is decoded by the selected neural decoder. 
We repeat the above procedure for all blocks. 
Next, we combine all blocks to reconstruct the entire image, by using the read shape information. 
Finally, the reconstructed image is passed through the deblocking network as a post-processing step.


\subsection{Training}

\textbf{Binarization Simulation}: The block-codes consist of floating-point numbers in the interval [0,1], which need to be binarized in order to achieve a reasonable compression rate.
Yet, binarization operation is non-differentiable and cannot be used as is for training the auto-encoder end-to-end.
Therefore, during training, we simulate the binarization by adding noise to a value $x[i]$ in the block-code $x$.
The noise is random with a uniform distribution within the interval $[-|nint(x[i])-x[i]|,|nint(x[i])-x[i]|]$ , where $nint$ denotes the rounding to nearest integer operation and $|.|$ is the absolute value operator.
The noise is selected such that the resulting value with additive noise remains in the interval $[0,1]$.
This is to be able to provide a consistent input to the decoder when we use this approximation and when we use the real binarization.

\textbf{Entropy-Friendly Loss}:
We concatenate the 1-dimensional block codes from the image, and the resulting image code is entropy-coded to achieve higher compression rates.
To make the image code more suitable for entropy coding, we propose the loss in Eq. \ref{entropyfriendly}.

\begin{equation}
\label{entropyfriendly}
L_{entropy}=\frac{1}{N-1} \sum\limits_{i=2}^{N}  (x_p[i]-x_p[i-1])^2
\end{equation}

In Eq. \ref{entropyfriendly}, $x_p$ is obtained by padding the code $x$ with 0 from both sides, where $x$ corresponds to a block-code and $N$ is the number of elements in $x_p$.
This padding is beneficial since in the end we will concatenate all the block codes, thus enforcing both beginning and ends of each block-code to be zero helps achieving a smooth image-code after concatenation.

\textbf{Training Process}: During training, we use MSE-based reconstruction loss $L_{rec}$ and the entropy loss with a regularization parameter $\lambda$ as follows: $L=L_{rec}+\lambda L_{entropy}$.

Although we simulate the binarization via additive random noise as previously discussed, we found it further beneficial to utilize an alternate training. 
In each epoch, we first train the auto-encoder end-to-end with binarization simulation over the entire training dataset.
Next, we freeze the encoder part, perform actual binarization on the codes and train only the decoder over the entire training dataset. 

Since we are going to use each of three neural networks for a different encoding difficulty level, the above training can be suboptimal. 
In fact, training for example the lowest bit rate neural network with all blocks (including the hardest blocks), would not be consistent with the inference stage, when that network would never been used on hard to encode blocks.
To make each network expert to their targeted blocks, we fine-tune each network with the blocks for which that network satisfies the target PSNR.

To make the decoder even more suitable to binarized codes, we keep the encoder frozen, use real binarization and fine-tune each expert decoder on its own training blocks.

The training of the deblocking network is performed separately where input images are the images reconstructed via the inference stage (except the deblocking part) and the ground truth are the original images. 

\section{Experimental Results}

In this section, we quantitatively evaluate our method in CLIC image compression dataset \cite{clicdataset}. 
As an evaluation metric we use the peak signal-to-noise ratio (PSNR).
We calculate a single mean-squared-error (MSE) from the entire image dataset and calculate the corresponding PSNR according to Eq. \ref{psnr}.
Note that the MSE is calculated on RGB images.

\begin{equation}
\label{psnr}
PSNR=20\log_{10}255 - 10\log_{10}MSE
\end{equation}

\subsection{Implementation Details}

All the convolutional or deconvolutional layers have 3x3 kernel size. The number of filters in the first five encoder's layers are 64, 128, 256, 512, 1024. The number of filters in the last layer of the encoder is different in each neural network: 64, 216 and 368 -- these determine the number of values output by each encoder. The decoder simply follows the filter sizes for encoder in reverse order. The final layer has 3 filters to convert back to RGB space. 
The training is performed on 32x32 blocks extracted from images from training dataset with half-overlapping blocks. 
We refer to the half-overlapping variant as data augmentation training. 
The regularization parameter for entropy coding loss is selected as $\lambda=0.001$. All neural networks were trained using a batch size of 256 and Adam optimizer with learning rate $0.001$.
We have two variants: \textit{NTcodec} where we apply deblocking filter on blocks of size 255x255 for memory efficiency, and \textit{NTcodecFull} where we apply deblocking filter on the entire image.

\begin{table*}
  \begin{center}
    \caption{Effect of Each Contribution for a Single Neural Network with 216 encoding dimension (on Validation Dataset)}
    \label{table1}
    \begin{tabular}{|c|c|c|c|c|c|c|c|c|} 
    \hline
      & \textbf{No Noise Sim.} & \textbf{No $L_2$} & \textbf{No Data Aug.} & \textbf{No Alt. Train.}  & \textbf{No Entropy Loss} &  \textbf{Batch-norm} & \textbf{Full Model}\\
      \hline
      \textbf{PSNR} & 23.882 & 26.778 & 26.946 & 25.751 & 27.258 & 26.882 & 27.055\\
      \hline
    \end{tabular}
  \end{center}
\end{table*}

\begin{table*}
  \begin{center}
    \caption{Effect of Additinal Operations (on Validation Dataset)}
    \label{table2}
    \begin{tabular}{|c|c|c|c|c|c|c|} 
    \hline
      &\textbf{Single NN} & \textbf{Multiple NN} & \textbf{Expert NN} & \textbf{Decoder Fine-Tune} & \textbf{Deblocking} & \textbf{Code Optimize}  \\
      \hline
      \textbf{PSNR} & 27.055 & 27.691 & 27.779 & 27.792 & 28.088 & 28.929  \\
      \hline
    \end{tabular}
  \end{center}
\end{table*}

\subsection{Effect of Each Contribution}

First, we investigate the effect of each contribution by controlled experiments. 
In particular, we investigate the effect of noise simulation, $L_2$ normalization, data augmentation, alternate training and entropy loss.
In each of these experiments, one of the above properties were removed and all others were kept fixed.
We also compare our full model with a standard architecture where after each convolution and deconvolution layer there is a batch-normalization layer.
In this standard model, we remove the introduced $L_2$ normalization layer, but keep all other components same.
We have conducted the experiments only on the 216-bit neural network. 
We report the obtained validation PSNR in Table \ref{table1}.

Noise simulation and alternate training have significant effects, as they are crucial for approximating the binarization process.
The network with $L_2$ normalization results into a compression rate of 0.151 bits per pixel (bpp), whereas the one without to 0.134 bpp.
The network without $L_2$ normalization can achieve the same performance (both in bpp and PSNR) with the network with $L_2$ normalization, however at encoding dimension of 236. 
As the encoding dimension increases, the training of the network takes longer, moreover the network size increases.
Therefore, $L_2$ normalization has a positive effect in training speed and final network size.
Data augmentation has a very minor effect on the performance due to already large number of training blocks and correlated blocks in the data augmentation.
The model with batch-normalization behaves similarly to no-normalization network in terms of bit-rate and PSNR, i.e. $L_2$ normalization achieves similar performance with faster training and lower number of network parameters.
Finally, since the entropy-loss acts as a regularization loss, it reduces the final PSNR value. 
However, the validation set bit rate (after entropy coding) with entropy-loss is 0.151 bits per pixel (bpp) whereas without the entropy-loss it is 0.216 bpp.
Therefore, the huge compression rate improvement dominates the slight PSNR decrease.

Next, we investigate the effect of multiple networks, expert neural network training, final decoder fine-tuning, code-optimization and deblocking post-processing.
Each experiment is done incrementally to each other in the above order.
We report the validation PSNRs and the bit rates for each incremental training in Table \ref{table2}.
As we observe, using multiple neural networks provide a decent performance increase whereas expert trainings and decoder fine-tuning has only a minor incremental effect. 
Deblocking post-processing was aimed to help achieving visually better quality images, yet we also observe that it increases the performance too.
Finally, block-wise code optimization greatly improves the performance and achieves a decent PSNR.
We would like to note here that the average bit rate for the final model with code optimization is 0.149 bpp, which is below our baseline with single network with no additional processing (0.151 bpp).

\textbf{Test-set results}: 
Table \ref{table3} reports PSNR and bit rates on the test-set for our method and for two traditional codecs (JPEG and BPG). 


\begin{table}
  \begin{center}
    \caption{Comparison on Test Set}
    \label{table3}
    \begin{tabular}{|c|c|c|c|c|} 
    \hline
      &\textbf{JPEG}  & \textbf{BPG} & \textbf{OURS}  \\
      \hline
      \textbf{PSNR} & 25.612  & 29.587  & 27.920  \\
      \hline
      \textbf{bpp} & 0.149 & 0.148 & 0.148 \\
      \hline
    \end{tabular}
  \end{center}
\end{table}

\section{Conclusion}
We have proposed an end-to-end block-based auto-encoder system for learned image compression.
We have evaluated each building block of our method and have shown that each building block contributes to the performance to a degree.
Our novel contributions $L_2$ normalization, concatenation-enabling entropy-friendly loss, expert neural network fine-tuning and code optimization greatly contribute to our final performance.

{\small
\bibliographystyle{ieee}
\bibliography{ntcodec}
}

\end{document}